\documentclass[10pt,letterpaper]{article}

\usepackage[T1]{fontenc}
\usepackage[utf8]{inputenc}
\usepackage{lmodern}
\usepackage{microtype}

\makeatletter
\@ifundefined{DisableLigatures}{}{%
  \DisableLigatures[f]{encoding=*,family=*}%
}
\makeatother

\usepackage[top=0.85in,left=2.75in,footskip=0.75in]{geometry}

\usepackage{amsmath,amssymb}
\usepackage{textcomp,marvosym}

\usepackage{changepage}
\raggedright
\usepackage[table]{xcolor}
\usepackage{array}
\newcolumntype{+}{!{\vrule width 2pt}}
\newlength\savedwidth

\usepackage{graphicx}
\usepackage{subcaption}
\usepackage{float}
\usepackage[section]{placeins}
\usepackage{flafter}

\usepackage[aboveskip=1pt,labelfont=bf,labelsep=period,
            justification=raggedright,singlelinecheck=off]{caption}

\usepackage{cite}
\usepackage{lastpage,fancyhdr}
\fancyheadoffset[L]{2.25in}
\fancyfootoffset[L]{2.25in}
\makeatletter
\renewcommand{\@biblabel}[1]{\quad#1.}
\makeatother

\usepackage{hyperref}
\usepackage{nameref}

\DeclareUnicodeCharacter{202F}{\,}

\begin{document}

\vspace*{0.2in}

\begin{flushleft}
{\Large
\textbf\newline{Automated classification of natural habitats using ground-level imagery} 
}
\newline
\\
Mahdis Tourian\textsuperscript{1,2*},
Sareh Rowlands\textsuperscript{1,2},
Remy Vandaele\textsuperscript{1,2},
Max Fancourt\textsuperscript{3},
Rebecca Mein\textsuperscript{3},
Hywel T.P.Williams\textsuperscript{1,2},
\\
\bigskip

\textbf{1}  Centre for Environmental Intelligence, University of Exeter, Exeter, UK
\\
\textbf{2} Department of Computer Science, Faculty of Environment, Science and Economy, University of Exeter, Exeter, UK
\\
\textbf{3} Natural England, York, UK
\\
\bigskip

%
%





* M.Tourian@exeter.ac.uk

\end{flushleft}
\section*{Abstract}
Accurate classification of terrestrial habitats is critical for biodiversity conservation, ecological monitoring, and land use planning. Several habitat classification schemes are in use, typically based on analysis of satellite imagery and validation by field ecologists. Here we present a methodology for classification of habitats based solely on ground-level imagery (photographs), offering improved validation and enhanced ability to classify habitats at scale (e.g. using imagery from citizen science). In collaboration with Natural England, a public sector organisation with responsibility for nature/biodiversity conservation in England, this study develops a classification system that applies deep learning to ground-level habitat photographs, categorizing each image into one of 18 distinct classes following the `Living England' framework. Images were pre-processed using resizing, normalization, and augmentation techniques, while resampling was used to balance classes in the training data and enhance model robustness. We developed and fine-tuned a custom deep learning classifier based on the DeepLabV3-ResNet101 architecture to assign a habitat class label to ground-level photographs. Using five-fold cross-validation, the model demonstrated strong overall performance across 18 habitat classes, with accuracy and F1-scores varying between classes. This approach supports robust, scalable habitat classification based on balanced and well-prepared training data. Across all folds, the model achieved a mean F1-score of 0.61, with some habitat classes such as Bare Soil, Silt and Peat (BSSP) and Bare Sand (BS) reaching values above 0.90. High performance was achieved for visually distinct habitats and lower performance for mixed or ambiguous classes. These findings demonstrate the potential of this novel approach for ecological monitoring. Ground-level imagery is easily obtained and accurate computational methods for habitat classification based on such data have many potential applications. To support use by practitioners we also provide a simple web application that allows classification of uploaded images using our model.


\section*{Introduction}

Habitat classification is an essential tool for biodiversity conservation, land management, and ecological monitoring. While natural ecosystems are complex and exhibit many idiosyncrasies, habitat classification frameworks provide a method for reducing this diversity into a small number of defined classes. While a `habitat' is typically defined with respect to a single species, the classes used by such frameworks refer to `habitat types' that support many species and are often defined by the combination of vegetation and other biotic/abiotic factors that support wildlife in a particular location. Each habitat type in a classification framework serves as a proxy indicator for the range of species present at a given location, as well as for key features of the local ecosystem such as soil type, moisture balance, and climatic conditions \cite{rodwell1991}.
Various habitat taxonomies exist. For example, the Phase 1 framework for the UK \cite{JNCC_Phase1_2006} divides terrestrial habitat types into ten categories (e.g. A: Woodland and scrub, B: Grassland and marsh, C: Tall herb and fen, D: Heathland, and so on). More recently, the Living England framework labels every part of England with one of 17 main habitat classes \cite{NE_LivingEngland_2024}.

Applying a given taxonomy using manual classification methods is labour-intensive, time-consuming, and subject to human error \cite{downes2009observerbias}. This has led to the adoption of automated techniques based on satellite imagery and remote sensing, enabling scalable, cost-effective, and repeatable habitat mapping across large and heterogeneous landscapes \cite{jongman2019automated}. However, satellite-based approaches still require ground-level validation, typically performed by trained ecologists, which demands substantial time and effort.
In this work, we explore whether ground-level imagery (e.g. photographs of habitats) can be used to automatically classify the habitat type present at a location using computer vision techniques. If successful, this approach offers several potential benefits: ground-level imagery is easily obtained, and accurate automated classification would enable high-resolution, consistent habitat mapping at scale, without requiring expert annotation.

Recent advancements in deep learning \cite{zhao2024review}, particularly in convolutional neural networks (CNNs), have made image-based ecological analysis increasingly feasible. Deep learning methods are already playing a growing role in ecological monitoring and ecosystem science,  supporting tasks such as wildlife detection from camera traps, acoustic species recognition, and environmental change monitoring from satellite imagery \cite{barta2024deep, perry2022outlook}. These tools are also advancing toward greater interpretability and causal understanding through hybrid models and attention mechanisms.

Numerous studies have explored computer vision methods for wildlife monitoring, and the field continues to evolve rapidly. For example, Miao et al. \cite{miao2019deep} used CNNs to classify African wildlife in camera-trap images, achieving high accuracy and employing interpretability techniques such as Grad-CAM (Gradient-weighted Class Activation Mapping) \cite{selvaraju2017grad} to highlight image regions influencing model predictions. Similarly, Otsuka et al. \cite{otsuka2024exploring} applied deep learning to classify seabird behaviours using accelerometer data, demonstrating the effectiveness of hybrid CNN–LSTM models combined with self-attention mechanisms. Bi et al. \cite{bi2024ecosystem} showed how deep learning can enhance ecosystem health assessments by integrating complex environmental datasets efficiently.

With regard to the specific problem of habitat type classification, recent advances have been made, though key challenges remain. 
In the marine domain, Game et al. \cite{game2024machine} proposed a hybrid CNN–SVM model to classify benthic habitats using imagery. Their results demonstrated the potential for simplified and practical deep learning pipelines in ecological research. 
Leblanc et al. \cite{leblanc2024deep} developed a method to classify terrestrial habitat types using the EUNIS framework. Rather than relying on imagery, their approach combined species composition data with environmental variables in a deep learning model to classify vegetation across Europe. The study underscored the importance of integrating biological and environmental data for robust ecological predictions. However, such detailed datasets are not always available or complete at national scales like the UK, particularly for fine-grained habitat mapping. This limitation motivates our exploration of whether ground-level imagery alone can support reliable habitat classification through computer vision.

Meanwhile, habitat classification approaches using satellite imagery are more common in terrestrial contexts. For instance, the Living England project \cite{NE_LivingEngland_2024} uses satellite data combined with machine learning algorithms to assign each 10m pixel in England to one of 17 single-type habitat classes and 1 mixed-type class. The approach emphasizes national-scale consistency and is updated regularly to support monitoring and land-use decision-making. Similarly, the Centre for Ecology \& Hydrology (CEH) Land Cover Maps \cite{CEH_LCM_2020} provide long-term, high-resolution classifications of UK land cover based on satellite imagery, with supervised classification techniques applied to multispectral data. These frameworks demonstrate the scalability and repeatability of satellite-based approaches, though they still require field validation and may struggle with fine-grained habitat type distinctions, especially in heterogeneous or transitional landscapes.

To our knowledge, no studies have yet explored the use of ground-level imagery for classifying terrestrial habitat types, a potentially valuable approach. Indeed, ground-level images can be easily acquired and shared by non experts using devices such as smartphones to be then collected through social media scrapping or citizen science initiatives. These images could then be classified by experts or algorithms to provide ground-level habitat information. Image-based habitat classification faces several challenges. Previous image-based ecological studies have reported frequent misclassifications in cases involving complex or visually similar habitats (for example, in wildlife identification tasks \cite{bothmann2023automated}), and noted difficulties in generalizing across geographic regions (such as in wetland mapping efforts \cite{mainali2023cnn}). Ethical and technical challenges such as data bias, interpretability, and model transferability must also be addressed \cite{barta2024deep,bi2024ecosystem}.


Here we present a deep learning approach to habitat type classification using ground-level images of UK habitats. The method is aligned with the Living England habitat taxonomy and is trained on photographs captured by ecologists across England as part of the Living England validation process. The dataset includes 18 distinct habitat types, based on the Natural England classification framework. We develop a classification model using a transfer learning strategy and evaluate its performance through cross-validation, analysing per-class accuracy and common misclassification patterns.
To support practical applications and user engagement, we also developed a simple web-based tool that enables users to upload images and receive habitat class predictions. This interface serves as a foundation for iterative model refinement and real-world usability. To our knowledge, no prior work has addressed habitat classification using ground-level imagery within the UK context, particularly one aligned with a national conservation framework such as Living England. Our study represents a first step toward adapting and evaluating computer vision approaches to habitat classification using ground-level imagery in the UK.


\section*{Methods}

Our experiment created a neural network classifier to automatically label images of habitats with a class from the Living England habitat type taxonomy. This is a challenging computer vision task since a `habitat' is an assemblage of co-occurring vegetation and biotic/abiotic factors rather than a single object. While habitat types can be visually distinctive, the classification depends on the simultaneous presence of multiple visual features including vegetation, colour, bare rocks/soil, etc. Our methodology is based on transfer learning with pre-trained vision models that have demonstrated skill in object classification. This section describes the training dataset of labelled habitat images, how the dataset was prepared for use,  the transfer learning approach and the pre-trained vision models used, and the experiment design and performance metrics. 

\subsection*{Dataset}

The dataset consists of 43,092 RGB-encoded ground-level photographs captured by ecologists working for Natural England as part of the Living England habitat mapping initiative \cite{NERR108}. Each image was taken by an ecologist during a field site visit, labelled with a habitat class from the Living England taxonomy, and linked with metadata on field site, location, and collection date.

Living England is a national-scale habitat mapping project covering the entirety of England up to Mean High Water Springs, first launched in 2016 as Defra’s "Living Maps" pilot and now on its fourth phase for 2021–22 data collection \cite{NERR108,Kilcoyne2022Blog}. It maps 18 broad habitat classes aligned with UK Biodiversity Action Plan categories (17 single-habitat classes along with an 18th {\it Multiple} class reflecting mixed habitats). A hybrid workflow uses computer vision to segment satellite imagery (Sentinel-1/2) at $\sim10m$ resolution and applies object-based random forest classification alongside targeted field surveys for model training and validation \cite{NERR108,DefraBlog2024}. The current version (2022--23) achieves approximately 87--88\% overall accuracy and is updated biennially under an Open Government Licence \cite{Trippier2024,Woodget2024}. 

Table ~\ref{tab:class_distribution}
shows the number of images available for each habitat class. The dataset is highly imbalanced, with certain categories such as {\it Improved and Semi-Improved Grassland} (IG) having over 10,000 images, while others like {\it Bare Soil, Silt and Peat} (BSSP) contain fewer than 300 images. Class imbalance is known to impact classification performance in computer vision and was addressed during model training through resampling (see below).

\begin{table}[htbp!]
\centering
\caption{Number of images per habitat category in the Living England dataset.}
\begin{tabular}{|l|c|r|}
\hline
\textbf{Category} & \textbf{Abbreviation} & \textbf{Number of Images} \\
\hline
Arable and Horticultural & AH & 2359 \\
Bare Sand & BS & 957 \\
Bare Soil, Silt and Peat & BSSP & 224 \\
Bog & BOG & 1750 \\
Bracken & BRA & 2567 \\
Broadleaved, Mixed and Yew Woodland & BMYW & 3187 \\
Built up areas and Gardens & BUAG & 754 \\
Coastal Saltmarsh & CS & 1008 \\
Coastal Sand Dunes & CSD & 1546 \\
Coniferous Woodland & CW & 371 \\
Dwarf Shrub Heath & DSH & 4699 \\
Fen, marsh and swamp & FMS & 2044 \\
Improved and Semi-Improved Grassland & IG & 10555 \\
Inland rock & IR & 794 \\
Multiple & Multiple & 1593 \\
Scrub & SCR & 2053 \\
Unimproved grassland & UG & 6172 \\
Water & WAT & 459 \\
\hline
\textbf{Total} &  & \textbf{43092} \\
\hline
\end{tabular}
\label{tab:class_distribution}
\end{table}

\begin{figure}[htbp!]
    \centering
    \includegraphics[width=\textwidth]{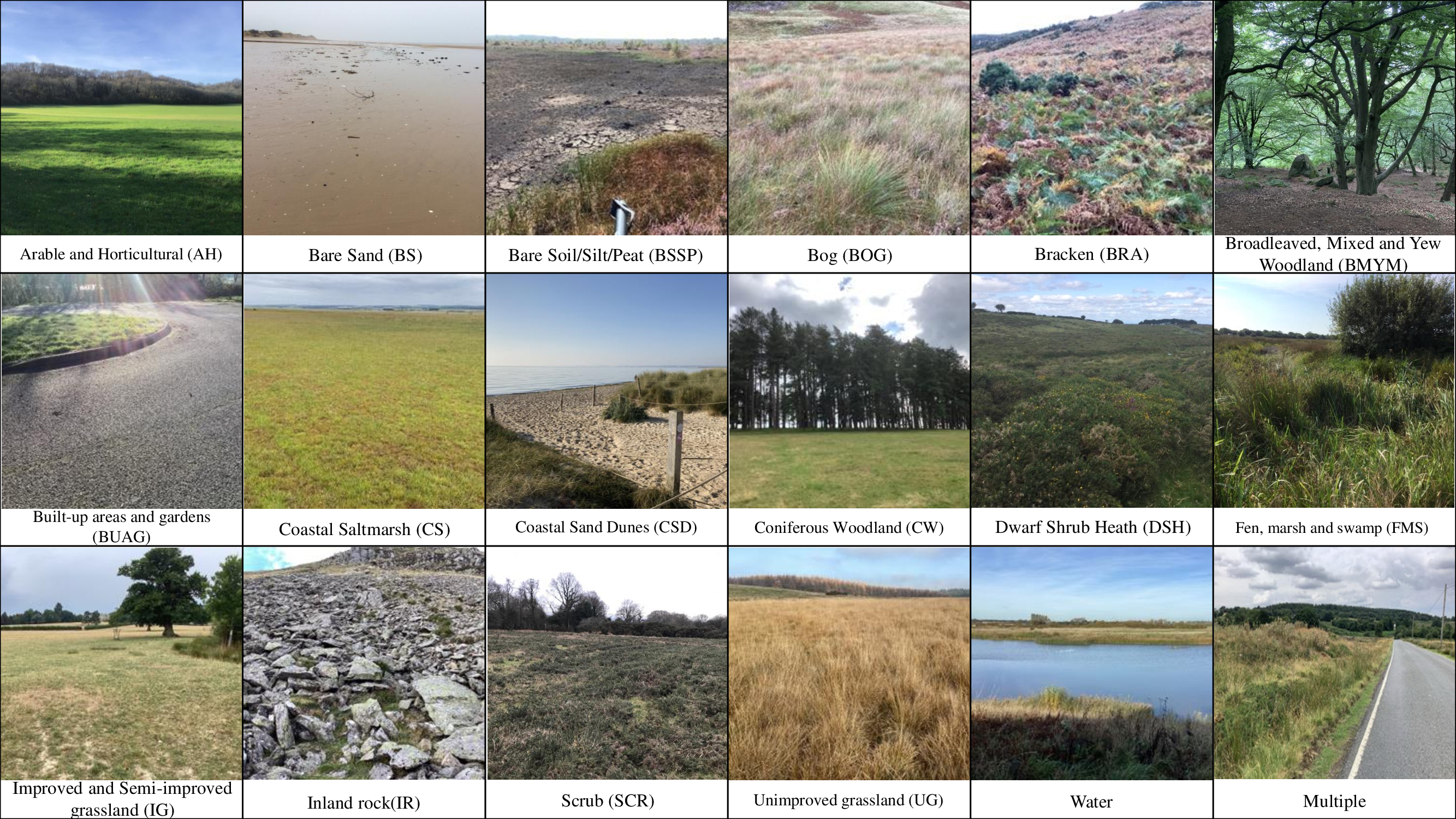}
    \caption{
Examples of the 18 habitat classes in the Natural England habitat classification taxonomy.
Images sourced from: Natural England (licensed under the Open Government Licence v3.0);
Copernicus Sentinel data © 2021–2023 (processed by the authors);
APGB imagery used under institutional access, subject to the APGB End User Licence.
}
    \label{fig:habitat_classes}
\end{figure}

Examples of images from each class are shown in Figure~\ref{fig:habitat_classes}. The {\it Multiple} class refers to images that contain visual characteristics from more than one habitat category (e.g., a transitional area between scrub and grassland, or a wetland with adjacent woodland). From our observations, these mixed-class images reflect ecotones or edge environments that do not clearly belong to a single dominant class. Although they represent a small part ($<4\%$) of the dataset, they introduce ambiguity that can challenge classification models.

%


\subsection*{Classifier Model \& Training}

We adopted a transfer learning approach by leveraging pre-trained deep convolutional neural networks (CNNs), which were fine-tuned for the habitat classification task using ecologist-captured imagery.
In preliminary work, we tested a standard image classification model based on InceptionV3 for the habitat classification task. This model was trained for 200 epochs with a custom classification head for habitat prediction. The InceptionV3-based model exhibited early convergence, with a peak validation accuracy of 44\% and high variance across epochs, indicating instability and limited learning capacity. 
We next tested a classifier based on DeepLabV3, which is pre-trained on outdoor scene segmentation and better suited to our habitat image dataset. This showed better performance and was chosen for further development.

The final model was based on DeepLabV3 with a ResNet-101 backbone, originally designed for semantic segmentation \cite{chen2017rethinking}. DeepLabV3 incorporates atrous convolutions and an Atrous Spatial Pyramid Pooling (ASPP) module to capture multi-scale context efficiently. We adapted this architecture for image-level classification by replacing the segmentation head with a 1×1 convolutional layer followed by global average pooling and flattening. This design leverages the spatially aware, high-resolution features of DeepLabV3 while outputting image-level predictions. Dropout (rate = 0.5) was applied for regularization. 

\subsection*{Data Preparation}

Since our chosen deep learning model, DeepLabV3–ResNet101, was pre-trained on the ImageNet dataset \cite{deng2009imagenet}, we followed a similar preprocessing scheme to ensure compatibility and effective transfer learning. All images were resized to 224×224 pixels and normalised using standard ImageNet statistics (mean = [0.485, 0.456, 0.406]; standard deviation = [0.229, 0.224, 0.225]). For the training set, we applied data augmentation techniques, including random horizontal flipping, $\pm$15° rotation, colour jittering, and the AutoAugment ImageNet policy \cite{cubuk2019autoaugment}, to improve generalisation and mitigate overfitting. Validation images were resized and normalised without augmentation.
To address class imbalance in the Natural England dataset, we standardised the number of training images per class to 1,000. For overrepresented classes, we randomly subsampled; for underrepresented classes, we synthetically expanded the dataset using the aforementioned augmentation strategy until 1,000 images were reached. This threshold was selected as a trade-off between computational feasibility and representation diversity, and aligns with common practice in image classification pipelines.
Training was accelerated using automatic mixed precision (AMP) \cite{micikevicius2018mixed}, which reduces memory usage and improves runtime efficiency without compromising model accuracy.

\subsection*{Experimental Setup}
\label{sec:training}

The model was trained using the Cross-Entropy Loss function and optimized with AdamW (learning rate = 1e-4, weight decay = 1e-4). Mixed precision training was employed using PyTorch AMP to accelerate training and reduce memory usage \cite{pytorch-amp}. To further minimize GPU memory consumption, gradient checkpointing (using {\it torch.utils.checkpoint}) was applied to both the backbone and classifier components of the network, allowing recomputation of intermediate activations during backpropagation. Training was conducted with a batch size of 16 for up to 100 epochs, with early stopping triggered if performance did not improve for seven consecutive epochs. 

Accuracy (see below) was used as the primary metric to monitor model performance during training. Specifically, validation accuracy was tracked across epochs, and the model with the highest validation accuracy was selected as the best-performing checkpoint for each fold. 
To validate our method, we used five-fold cross-validation: the data set was divided into five folds, and the model was trained and evaluated five times, each time using a different fold as the validation set and the remaining folds for training. 

The validation images were transformed through a simpler transformation pipeline without augmentations to ensure consistent evaluation. This approach is standard in machine learning: data augmentations (e.g., random cropping, flipping, color jitter) are applied during training to improve generalisation by exposing the model to a wider variety of inputs. However, such augmentations are not applied to validation images because the goal during validation is to evaluate model performance on data that resembles real-world, unseen inputs as closely as possible. Applying augmentations during validation could introduce artificial variation and lead to misleading performance estimates.

\subsection*{Evaluation Metrics}

Model performance was evaluated using several standard classification metrics: accuracy, precision, recall, and F1-score, calculated on a per-class basis. 
Metrics were defined as follows:

\begin{equation}
\text{Accuracy} = \frac{\text{Number of Correct Predictions}}{\text{Total Number of Predictions}} = \frac{TP + TN}{TP + TN + FP + FN}
\end{equation}

\begin{equation}
\text{Precision} = \frac{TP}{TP + FP}
\end{equation}

\begin{equation}
\text{Recall} = \frac{TP}{TP + FN}
\end{equation}


\begin{equation}
\text{F1} = 2 \times \frac{\text{Precision} \times \text{Recall}}{\text{Precision} + \text{Recall}}
\end{equation}

\noindent where TP, TN, FP, and FN refer to true positives, true negatives, false positives, and false negatives, respectively. We calculated these values separately for each class using a one-vs-rest approach, where each class is treated as positive and all others as negative. Precision shows how many predicted positives were correct, recall shows how many actual positives were found, and the F1 Score combines both to give an overall measure of performance\cite{sokolova2009systematic}. 
Confusion matrices were used to visualise misclassification patterns. 

 In addition, we use categorical cross-entropy loss to evaluate model performance on the validation set after each training epoch. This loss measures the difference between the predicted class probabilities and the true class labels. It is defined as:

\begin{equation}
\mathcal{L} = -\sum_{i=1}^{C} y_i \log(\hat{y}_i)
\end{equation}
where \( C \) is the number of classes, \( y_i \) is the true label, and \( \hat{y}_i \) is the predicted probability for class \( i \). Validation loss is used both to monitor generalisation and as the criterion for early stopping.

To ensure stability, metrics were computed across cross-validation folds and aggregated for analysis. 
Furthermore, we recorded metadata for each test image across all folds, including the predicted and true labels, class probabilities, and the top-3 most likely habitat classes predicted by the model. This enriched metadata enables deeper analysis of model uncertainty, for example, instances where the correct label was not the top-1 prediction but still appeared within the top-3 candidates. 

\section*{Results}

As discussed, we initially tested a standard classification model using InceptionV3, which achieved a peak validation accuracy of 44\% , its learning was unstable and prone to early convergence. The averaged precision, recall, and F1-score were 0.40, 0.38, and 0.39, respectively, reflecting its limited ability to generalize across habitat classes.
Based on these limitations, we explored DeepLabV3-ResNet101, a model originally designed for semantic segmentation but adapted here for image-level classification. The DeepLab-based model achieved significantly higher performance, with a higher average validation accuracy of 0.61, and averaged precision, recall, and F1-score of 0.63, 0.61, and 0.61, respectively. 
This improvement highlights the benefit of architectures that can better capture and use spatial information in an image for example, by looking at larger areas at once, combining features at different scales, or learning how different parts of the image relate to each other. This kind of model is especially useful for complex scene classification tasks like habitat recognition, where details such as vegetation patterns and background landscape provide important context.

\begin{figure}[htbp!]
    \centering
    \includegraphics[width=0.85\textwidth]{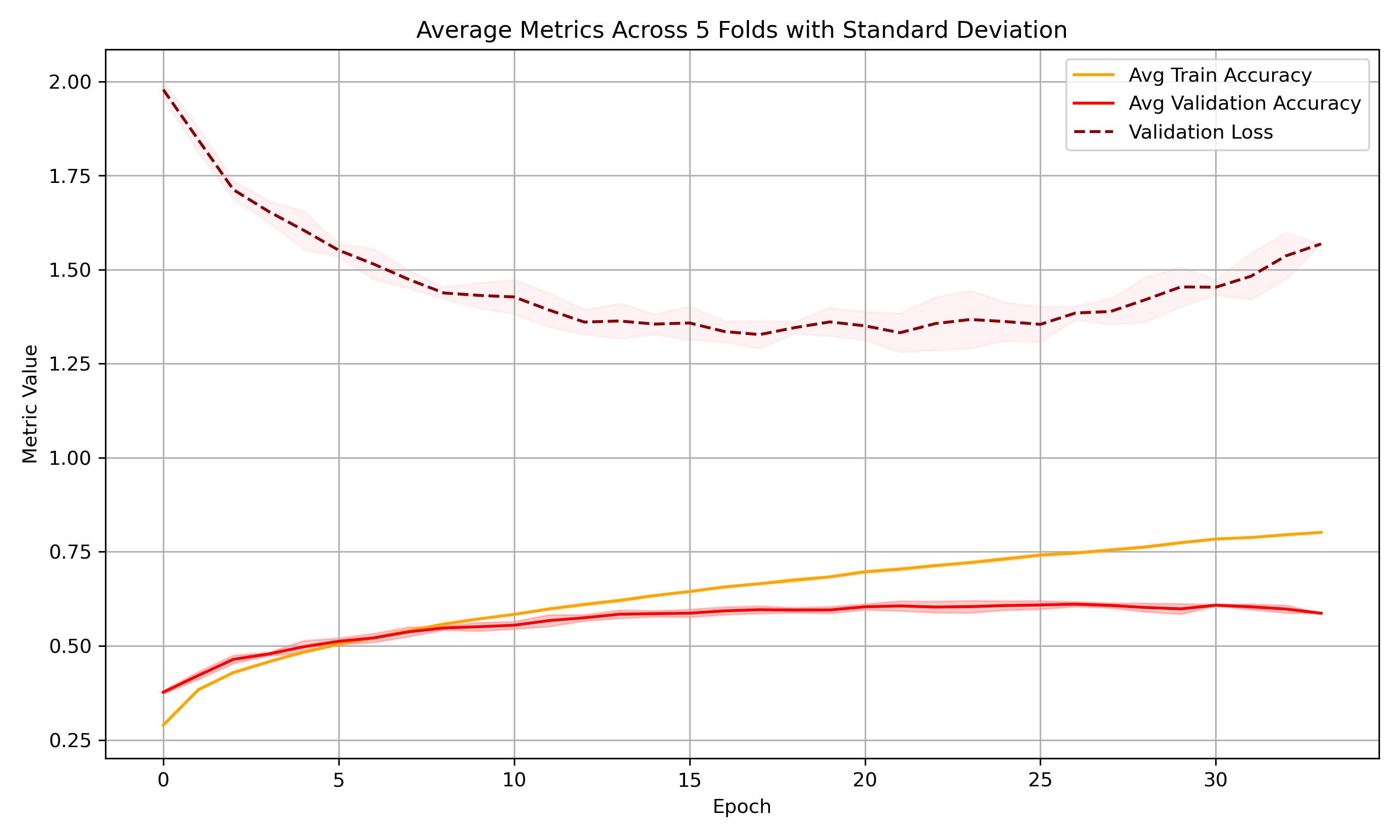}
    \caption{Average training accuracy, validation accuracy, and validation loss across 5-fold cross-validation, with shaded error bands representing the standard deviation at each epoch. Training accuracy steadily increases, while validation accuracy plateaus after around epoch 15, suggesting the onset of overfitting. Validation loss mirrors this trend, decreasing initially but rising again in later epochs.
    Early stopping occurred at epochs 29, 33, 30, 34, and 28 for folds 1 through 5, respectively.}
    \label{fig:avg_metrics}
\end{figure}

The DeepLabV3-ResNet101 model demonstrated steady learning across folds, with validation accuracy ranging from approximately 59\% to 63\% before early stopping. Fig~\ref{fig:avg_metrics} shows the average accuracy metrics across all folds. The model achieved consistent improvements in training and validation accuracy during the initial epochs, with validation performance plateauing around 63\%. While training accuracy continued to rise, indicating some overfitting, we applied early stopping based on validation accuracy. Specifically, training was halted if validation performance did not improve for a fixed number of epochs (patience=7), and the model parameters from the best-performing epoch were retained. This strategy helped mitigate overfitting and ensured that evaluation was based on the most generalizable version of the model.


\begin{table}[!ht]
\caption{{\bf Per-class performance metrics averaged across 5 folds.}}
\centering
\begin{tabular}{lccc}
\hline
\textbf{Class} & \textbf{Precision} & \textbf{Recall} & \textbf{F1-Score} \\
\hline
AH & 0.63 & 0.64 & 0.63 \\
BMYW & 0.60 & 0.61 & 0.61 \\
BOG & 0.46 & 0.56 & 0.50 \\
BRA & 0.67 & 0.74 & 0.70 \\
BS & 0.86 & 0.91 & 0.88 \\
BSSP & 0.88 & 0.94 & 0.91 \\
BUAG & 0.82 & 0.84 & 0.83 \\
CS & 0.55 & 0.60 & 0.57 \\
CSD & 0.53 & 0.51 & 0.52 \\
CW & 0.87 & 0.89 & 0.88 \\
DSH & 0.49 & 0.52 & 0.50 \\
FMS & 0.40 & 0.38 & 0.39 \\
IG & 0.43 & 0.51 & 0.47 \\
IR & 0.79 & 0.77 & 0.78 \\
Multiple & 0.29 & 0.18 & 0.22 \\
SCR & 0.50 & 0.42 & 0.46 \\
UG & 0.36 & 0.21 & 0.26 \\
WAT & 0.86 & 0.86 & 0.86 \\
\hline
\end{tabular}
\label{tab:per_class_metrics}
\end{table}

\begin{figure}[htbp]
    \centering
    \includegraphics[width=\textwidth]{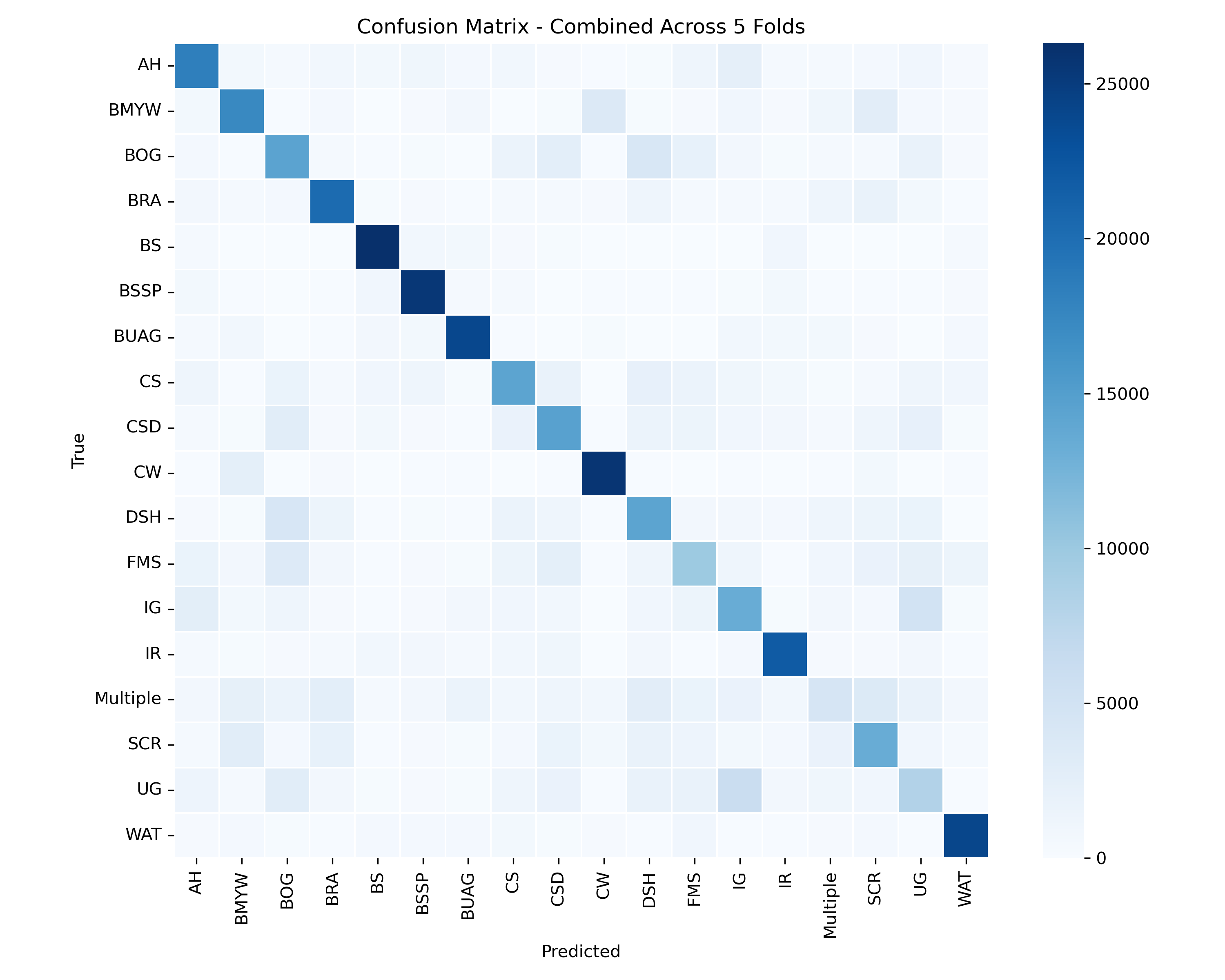}
    \caption{Confusion matrix showing class-wise prediction outcomes (summed across five folds). Diagonal values indicate correct predictions, while off-diagonal values show evidence of systematic prediction errors for some pairs of habitat types.}
    \label{fig:conf_matrix}
\end{figure}

Table~\ref{tab:per_class_metrics} summarizes the average precision, recall, and F1-score for all 18 classes. 
Fig~\ref{fig:conf_matrix} shows classification accuracy and also reveals patterns of mis-classification. The per-class performance varied considerably. 
Inspection showed that classes with well-defined visual features and less intra-class variation (e.g., {\it Water} (WAT), {\it Bare Sand} (BS), {\it Bare Soil, Silt and Peat} (BSSP)) performed better, typically exhibiting strong diagonal dominance in confusion matrices and F1-scores above 0.85. In contrast, ambiguous or visually mixed classes such as {\it Multiple} and {\it Unimproved Grassland} (UG) were often confused with others, as seen in their low recall and precision values.
Examining classification performance more closely, 
even well-performing classes are subject to some misclassification. 

\begin{figure}[htbp]
    \centering
    \includegraphics[width=0.7\textwidth]{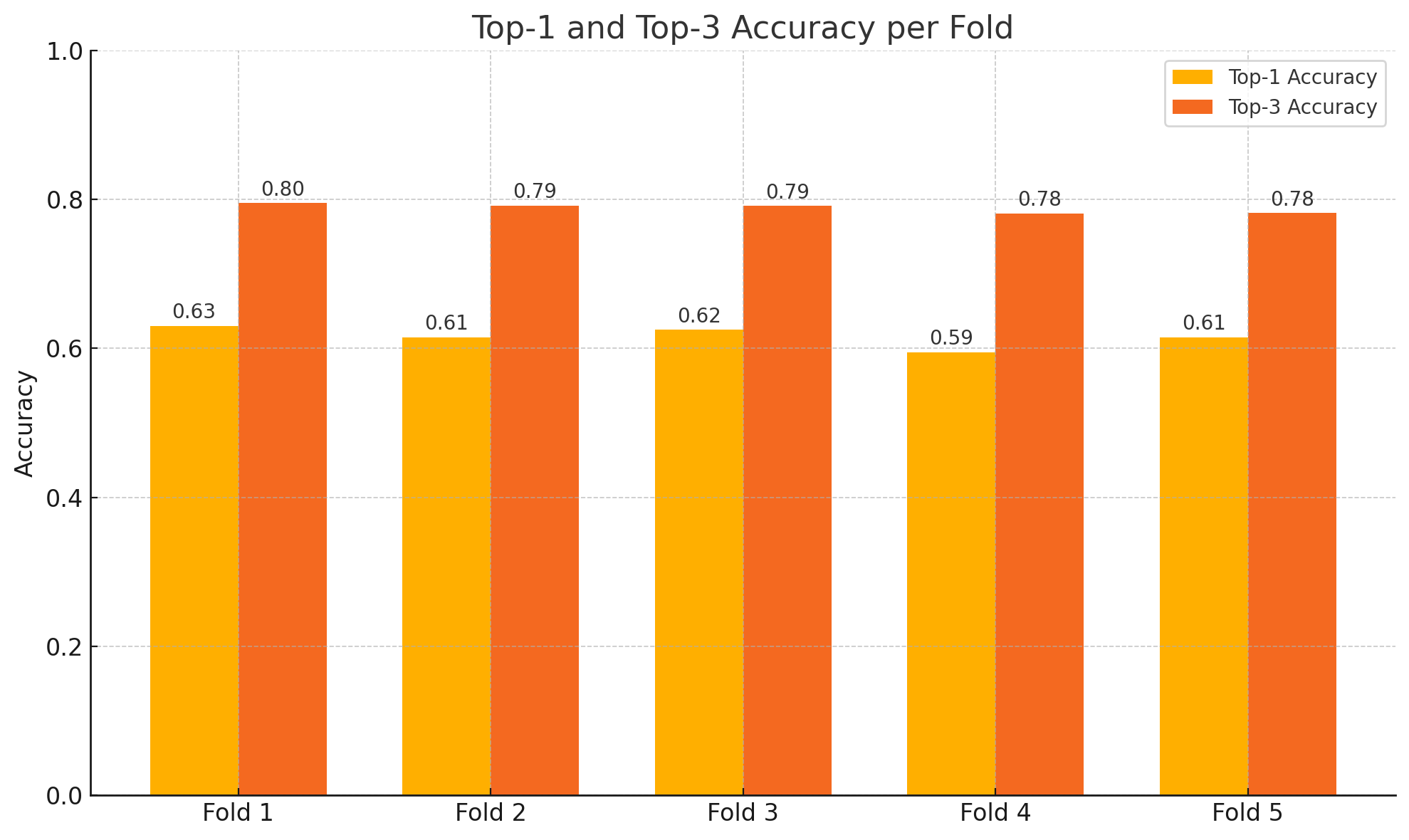}
    \caption{Top-1 vs. Top-3 classification accuracy across five cross-validation folds. While Top-1 accuracy remains consistent around 60--63\%, considering the Top-3 predictions significantly improves accuracy to approximately 78--80\%, indicating the model often ranks the correct class among its top predictions.}

    \label{fig:top1-top3_folds}
\end{figure}

To further understand model behaviour, we analysed stored metadata including predicted and true labels, top-3 predictions, and confidence scores. Figure~\ref{fig:top1-top3_folds} visualizes the frequency with which the true label appeared in the top-1 or top-3 predicted labels across the five cross-validation folds. The Top-1 accuracy reflects the proportion of predictions where the model's first choice matches the true label. Top-3 accuracy indicates cases where the true label appears among the top three predicted classes, offering insight into the model's performance when some uncertainty is allowed. The overall Top-1 classification accuracy across all five folds ranged from 59.0\% to 63.0\%, while Top-3 accuracy varied between 78.0\% and 80.1\%. This indicates that even when the Top-1 prediction was incorrect, the true label often appeared among the Top-3 predictions, especially for visually ambiguous classes such as \textit{Multiple} and \textit{UG}. This suggests that Top-3 confidence rankings retain ecologically useful information and could support soft-label evaluation or interactive human-in-the-loop feedback systems.

Close examination of the confusion matrices in Fig~\ref{fig:conf_matrix} allows analysis of misclassification patterns.
For instance, {\it Bare Sand} (BS) was most frequently confused with {\it Bare Soil/Silt/Peat} (BSSP) and {\it Coastal Sand Dunes} (CSD), which likely stems from similar sandy textures and coloration. {\it Water} (WAT), while generally distinctive, was occasionally misclassified as {\it Bare Sand} (BS) and {\it Coastal Saltmarsh} (CS), possibly due to shared visual features such as water reflections, wet soil, or adjacent vegetation. {\it Coniferous Woodland} (CW) was most often confused with {\it Broadleaved, Mixed, and Yew Woodland} (BMYW), a reasonable outcome considering the structural and chromatic overlap between coniferous and mixed woodlands in certain views. {\it Bare Soil/Silt/Peat} (BSSP) showed minimal confusion overall but was occasionally misidentified as {\it Bare Sand} (BS) and {\it Arable and Horticultural} (AH), both of which can exhibit bare ground or soil-like surfaces. These confusion patterns suggest that while some habitat types are easily distinguishable, others remain visually ambiguous due to ecological or spatial similarities. 

Some classes, particularly {\it Multiple} and {\it Unimproved Grassland} (UG), consistently showed lower classification performance, as reflected in both precision and recall scores. This can be partly explained by their inherent visual variability and substantial overlap with other categories. For example, Fig~\ref{fig:misclassification} shows that images labelled as {\it Multiple} often include a mix of habitat elements, making it visually similar to structured vegetation classes like {\it Scrub} (SCR). Similarly, {\it Unimproved Grassland} (UG) is frequently confused with {\it Dwarf Shrub Heath} (DSH) due to similar ground textures and vegetation patterns. 

\noindent\begin{minipage}{\linewidth}
  \centering
  \includegraphics[width=0.85\textwidth]{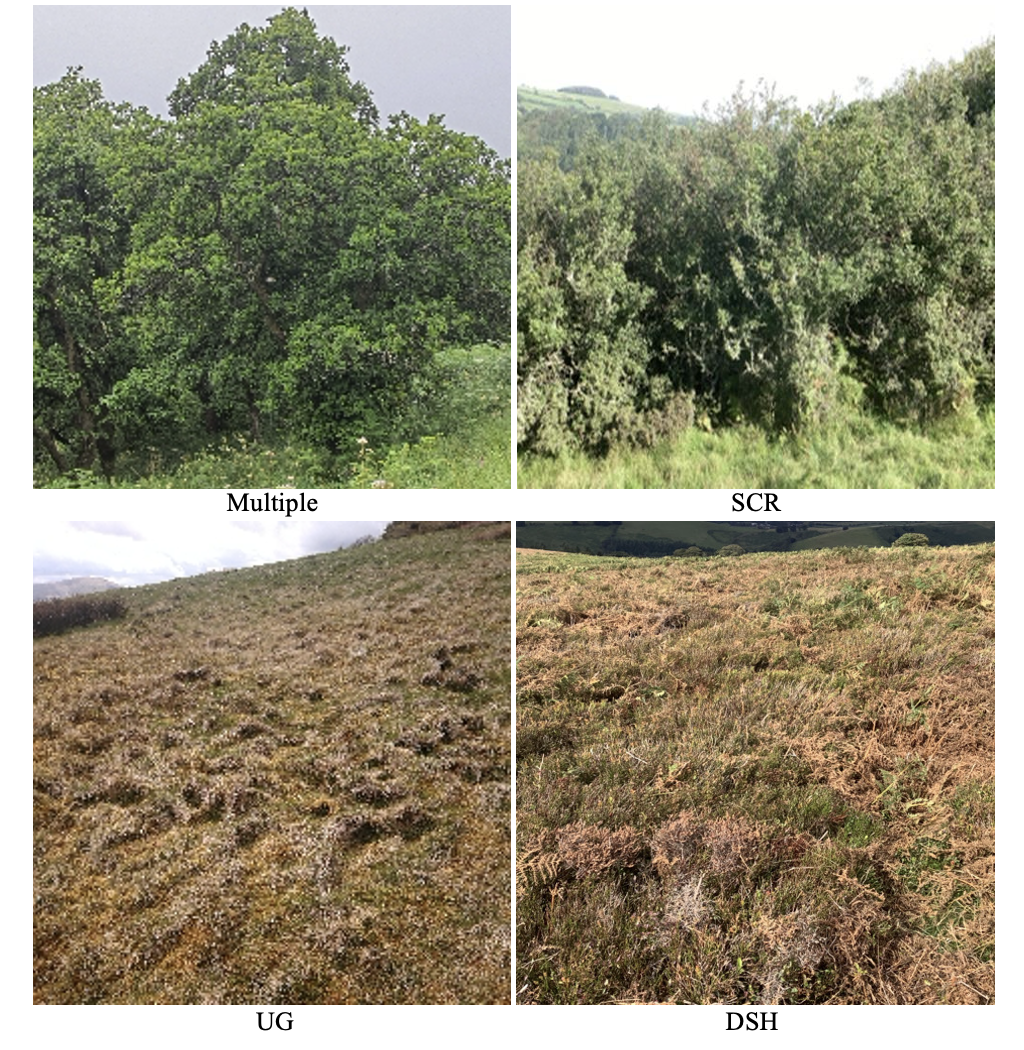}

  \captionsetup{belowskip=0.6\baselineskip}
  \captionsetup{hypcap=false}
  \captionof{figure}{Examples of misclassifications due to visual similarity between habitat classes. 
\textbf{Top row:} An image from the {\it Multiple} class (top-left) was misclassified as {\it Scrub} (SCR, top-right). Both exhibit dense green vegetation and overlapping structural features, likely contributing to the confusion. 
\textbf{Bottom row:} A {\it Unimproved Grassland} (UG) image (bottom-left) was predicted as {\it Dwarf Shrub Heath} (DSH, bottom-right). These habitats share similar visual elements, such as low vegetation and heterogeneous textures, which may explain the frequent misclassifications between them.}
  \label{fig:misclassification}
\end{minipage}

These examples suggest that some classification errors might stem not from model weaknesses but from genuine ecological or visual ambiguities, underscoring the need for either additional metadata or multi-label classification approaches in future work.

\begin{figure}[htbp]
    \centering
    \includegraphics[width=0.85\textwidth]{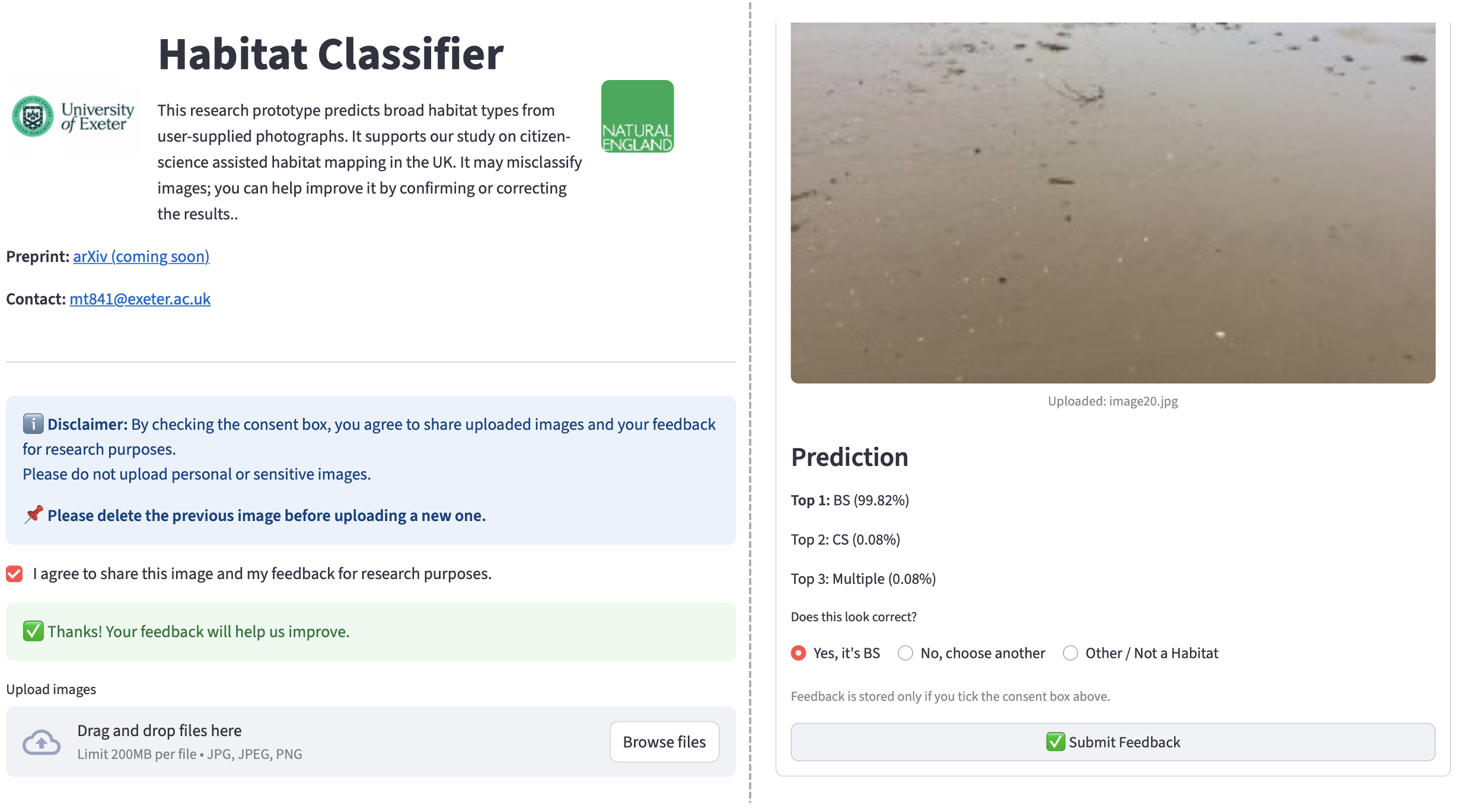}
    \caption{The landing page of the Habitat Classifier app, allowing users to upload ground-level habitat images for classification. An example of prediction results from the app showing top-3 habitat class predictions for an uploaded image.}
    \label{fig:app_results}
\end{figure}

To facilitate practical deployment and public engagement, we developed a web-based application for habitat classification. The app is available at \href{http://habitat-classification-web-app.duckdns.org/}{Habitat Classification Web App} and is built using Streamlit, a lightweight Python framework for interactive data apps. The interface allows users to upload one or more ground-level habitat images (in \texttt{.jpg}, \texttt{.jpeg}, or \texttt{.png} format), which are processed through a pre-trained DeepLabV3-ResNet101 classifier. Images are resized to 224$\times$224 pixels and normalized using ImageNet statistics prior to inference. The model outputs the top-3 predicted habitat classes along with their associated probabilities and definitions. 
In addition to providing predictions, the updated application supports user feedback collection to improve future model iterations. Users are prompted to confirm or correct the predicted label, or to specify a custom label if the result is inaccurate or ambiguous. These interactions are logged, along with the uploaded image and prediction confidence, into structured \texttt{.csv} and \texttt{.json} files. With user consent, submitted images and annotations are saved locally to support downstream research and dataset expansion.
The application also includes basic safeguards such as a consent checkbox and reminders not to upload personal or sensitive content. It serves as a prototype for a future mobile tool designed to support ecological monitoring in the field, targeting both professionals and citizen scientists. Screenshots of the interface are shown in Fig~\ref{fig:app_results}.

\section*{Discussion}

In this study, we developed an automated visual classification system for identifying UK habitat types from ground-level photographs. Using a modified DeepLabV3-ResNet101 architecture and a balanced dataset of 18 habitat classes, we demonstrated that accurate image-based classification is feasible even for complex ecological scenes. When seeking to predict a single correct habitat type for an image, the models achieved a mean F1-score of 0.61 across five-fold cross-validation, with high performance for visually distinct classes such as {\it Bare Sand} (BS), {\it Bare Soil/Silt/Peat} (BSSP) and {\it Water} (WAT), and lower performance for visually mixed or ambiguous classes like {\it Multiple} and {\it Unimproved Grassland} (UG). For a more relaxed task of identifying the correct habitat type within the top-3 most likely predicted labels, performance was increased with a mean F1-score of 0.79. 

Our study demonstrates that fine-tuned convolutional neural networks (CNNs) can effectively classify terrestrial habitats from ecologist-captured, ground-level imagery. The models demonstrated consistent performance across most habitat categories, achieving moderate to high per-class accuracy using RGB data alone, particularly for classes with distinct visual characteristics.
Classification outcomes varied by habitat type. Classes with strong visual features, such as water bodies (WAT) or built-up areas (BUAG), achieved the highest precision and recall, suggesting that CNNs are well suited to detecting visually distinct habitats. In contrast, ambiguous or transitional categories, such as {\it Multiple} or {\it Unimproved Grassland} (UG), exhibited lower performance, with models frequently confusing them with neighboring classes. These challenges reflect broader issues in land cover classification, where visually similar habitats often produce overlapping features and annotation difficulties \cite{mainali2023cnn,bothmann2023automated}.

The patterns observed in this study align with findings from related domains. For example, wildlife image classification and marine habitat detection have similarly shown CNNs to be effective for structured ecological imagery \cite{miao2019deep},\cite{game2024machine} . However, like in those fields, performance drops were observed for classes with greater within-class variation or unclear boundaries. This reinforces known limitations of deep CNNs when applied to complex natural imagery without supporting metadata.
To improve classification for visually ambiguous classes, future work could explore integrating additional metadata, such as GPS coordinates, elevation, soil type, and seasonal indicators. These contextual cues may help disambiguate classes with overlapping visual characteristics and support more robust predictions across ecological gradients.

Model enhancements through architectural innovations also offer promising directions. Transformer-based vision models (e.g., Vision Transformers \cite{dosovitskiy2021vit})
 may improve classification by capturing global spatial dependencies and long-range contextual features \cite{dosovitskiy2021vit}. This could be particularly useful for habitats with subtle texture differences across spatial scales. Ensemble learning approaches may also reduce variance and improve generalisation by combining complementary classifiers \cite{zhou2012ensemble}, especially when individual models overfit specific classes or data distributions. Few-shot learning methods may further aid underrepresented or difficult classes like \textit{Multiple} by enabling models to learn from limited examples \cite{wang2020generalizing}. These techniques have already demonstrated success in biodiversity and medical imaging contexts.

Accurate habitat classification contributes to broader ecological goals, including biodiversity assessment, conservation planning, and environmental monitoring. As taxonomic standards and classification schemes vary across regions and institutions, developing models that can generalize or be translated between different habitat taxonomies is critical. Aligning model outputs with standard classification frameworks (e.g., EUNIS, CORINE, or Living England) would facilitate broader integration with existing ecological datasets and policy frameworks.

Finally, to improve accessibility and practical impact, we deployed a web-based application that allows users to upload habitat images and receive top-3 model predictions. This app serves as a prototype for an eventual mobile tool that can support real-time habitat classification in the field. The intention is that this tool could be deployed as a mobile phone application, enabling users to capture habitat images and receive real-time habitat-type predictions. This tool would increase accessibility and support large-scale, crowd-sourced ecological monitoring efforts across diverse environments.

\section*{Acknowledgments}

We thank Natural England for providing access to the dataset used in this study.
This work was supported by UKRI EPSRC Grant No. EP/Y028392/1: AI for Collective Intelligence (AI4CI).

%
%
%

\bibliography{main}





\end{document}